\documentclass{article}
\usepackage{iclr2027_conference,times}

\usepackage{amsmath}
\usepackage{amssymb}
\usepackage{amsthm}
\usepackage{capt-of}

\usepackage{amsmath,amsfonts,bm}

\def\eqref#1{equation~\ref{#1}}

\def\1{\bm{1}}

\DeclareMathAlphabet{\mathsfit}{\encodingdefault}{\sfdefault}{m}{sl}
\SetMathAlphabet{\mathsfit}{bold}{\encodingdefault}{\sfdefault}{bx}{n}

\usepackage[breaklinks,colorlinks]{hyperref}
\hypersetup{
    colorlinks=true,
    linkcolor=red,
    citecolor=blue,
    filecolor=black,
    pdfstartview=Fit,
    breaklinks=true
}
\usepackage{url}
\usepackage{graphicx}
\usepackage{booktabs}
\usepackage{multirow}
\usepackage{xcolor}
\usepackage{colortbl}
\usepackage{xspace}
\usepackage{enumitem}
\newcommand{\method}{\textsc{Lethe}\xspace}

\newtheorem{proposition}{Proposition}

\title{LETHE: Defending against Unauthorized Image-to-Video Generation via Value Diversity Collapse}

\author{
Boheng Li$^{1}$, Yueming Xing$^{2}$, Yiming Li$^{1}$\thanks{Corresponding author. E-mail: \texttt{liyiming.tech@gmail.com}}, Jie Zhang$^{3}$, Tianwei Zhang$^{1}$ \\
$^{1}$Nanyang Technological University \\
$^{2}$Wuhan University  \quad
$^{3}$SiliconProspect AI
}

\iclrfinalcopy

\begin{document}

\maketitle

\begin{abstract}
Image-to-video (I2V) diffusion models can animate a single reference image into high-quality, subject-consistent videos, posing threats of misuse as anyone can turn personal photos into fabricated videos without consent. A promising solution is to proactively add an adversarial perturbation to the image before release, such that videos generated from it would be unfaithful or visually corrupted. However, due to an insufficient understanding of what governs how subsequent frames read the reference image in modern I2V models, existing protections struggle to offer strong protection under common perturbation budgets. In this paper, we approach the problem by exploring this underlying reading mechanism. Through a convex geometric analysis, we theoretically identify the spread of the reference value vectors as a key factor governing this reading process, and empirically find that even a mild contraction of this spread can severely degrade the generated video, making it an efficient lever for protection. Guided by this insight, we propose \method, a novel protection method that directly disrupts this reading process. \method pairs two complementary objectives, where a value collapse objective drives the reference value vectors toward uniformity, while an attention attraction objective draws subsequent frames toward these collapsed tokens to amplify corruption. We further construct two human-centric I2V protection benchmarks and conduct extensive experiments with multiple modern I2V models, which demonstrate the state-of-the-art protection performance and versatility of \method.
\end{abstract}

\section{Introduction}
\label{sec:intro}

Image-to-video (I2V) diffusion models~\citep{yang2025cogvideox,wan2025wan} have achieved remarkable progress in video synthesis. Given a reference image (e.g., a selfie) and a text prompt, these models can generate a high-quality video in which the subject faithfully retains the appearance in the reference image while performing the actions described in the prompt~\citep{huang2024vbench}. However, this capability also enables malicious individuals to animate a personal photo shared online into videos without the owner's consent, such as videos that fabricate the person's words and actions. Since such videos depict the person in natural motion, they can be deceptively realistic and thus pose serious threats to personal privacy and reputation, which have drawn growing public concern and legislative attention~\citep{gui2025i2vguard,song2026antiprompt,kamachee2025video}.

To mitigate this threat, one emerging solution is \emph{protective perturbation}~\citep{zhang2026adversarial}, which turns the vulnerability of deep learning models to adversarial perturbations~\citep{szegedy2013intriguing} into a means of protecting personal images. The key idea is to proactively add a small, carefully crafted perturbation to the photo before sharing it online, so that diffusion models taking the protected photo as input are misled into producing unfaithful or low-quality outputs~\citep{salman2023photoguard,liang2023advdm,xue2024toward,gui2025i2vguard,song2026antiprompt}. In this way, forgeries could be blocked at the source, substantially reducing the potential harm from their spread.

\begin{figure}[t]
\centering
\includegraphics[width=0.98\linewidth]{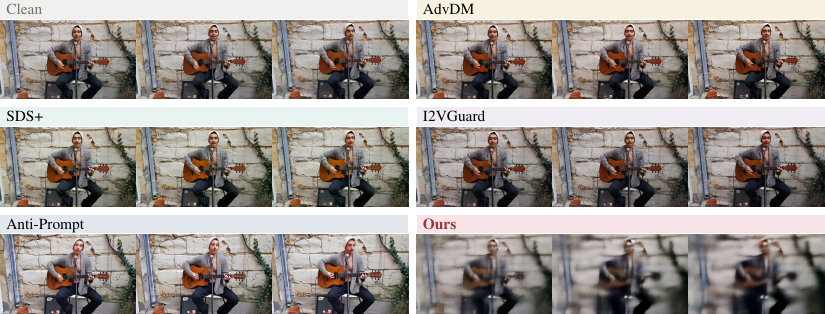}
\vspace{-0.5em}
\caption{Frames sampled from videos generated from reference images protected by different
methods. The perturbation budget is $16/255$ and the model is Wan2.1-I2V-14B~\citep{wan2025wan}.}
\label{fig:teaser}
\vspace{-2em}
\end{figure}

While protective perturbations have achieved notable success in image editing and customization~\citep{liang2023advdm,xue2024toward,salman2023photoguard}, protection against I2V generation is still at an early stage. Early efforts mainly targeted UNet-based models extended from Stable Diffusion, such as pose-driven animation models~\citep{zhou2025dormant}. Only recently, a few studies have begun to consider DiT-based models~\citep{long2026immune2v,song2026antiprompt}, the mainstream architecture of state-of-the-art I2V generation~\citep{shao2026efficient}. However, disrupting modern DiT-based video diffusion models appears considerably more challenging: as shown in Fig.~\ref{fig:teaser}, under the commonly used budget of $\epsilon=16/255$, which typically suffices to effectively disrupt image generative models~\citep{salman2023photoguard,xue2024toward}, the protection offered by existing methods is still not satisfactorily strong: they tend to introduce only local artifacts or mild visual changes into the generated videos, while the identity and appearance of the reference image remain somewhat recognizable. This raises concerns about the effectiveness of current protections and highlights the need for stronger protection tailored to modern attention-based I2V models.

To approach this problem, in this paper, we first seek to understand the key factor that governs how I2V models read information from the reference image. In modern I2V models, the reference image is encoded into a set of tokens, from which subsequent frames read primarily through attention operations~\citep{liu2026rolling}. Characterizing this reading process from a convex geometric perspective, we theoretically identify the spread of the reference value vectors, measured by the radius of their minimum enclosing ball, as an upper bound on the diversity of the reference readouts. This analysis points to reference-value diversity as a key target towards disrupting the reading process.  We further empirically find that even a mild contraction of this spread suffices to severely degrade the generated video, making it an efficient lever for protection. Motivated by this insight, we propose \method, which crafts protective perturbation by disrupting this reading process. \method pairs two complementary objectives, where the value collapse objective drives the reference value vectors toward uniformity, aiming to tighten the above bound, while another attention attraction objective draws the attention of subsequent frames toward these collapsed tokens to amplify disruption. In this way, \method constrains both the content and the source of the information read by subsequent frames, so that the protected photo can hardly be faithfully animated into a high-quality video. Lastly, since a standardized and publicly available benchmark tailored to human-centric I2V protection is still largely lacking, we construct two high-quality I2V protection benchmarks by re-annotating CelebV-Text~\citep{yu2023celebv} and Kinetics-700~\citep{carreira2019kinetics} with carefully curated reference frames and human-verified motion prompts. Extensive experiments on these benchmarks with multiple I2V models confirm the state-of-the-art effectiveness and versatility of \method.

In conclusion, our contributions are three-fold:
\begin{itemize}[leftmargin=*, itemsep=3pt, topsep=0pt, parsep=0pt, partopsep=0pt]
    \item We characterize the reference-reading pathway in attention-based I2V models from a convex geometric perspective and identify reference-value diversity as a key quantity that bounds the diversity of the reference readouts. Combining this analysis with direct value interventions, we establish reference-value diversity as a useful and efficient target for protective perturbations.
    \item We propose \method, a novel protection method that safeguards images against unauthorized I2V, by directly disrupting how these models read the reference image. With a pair of complementary objectives, \method jointly constrains the content and the source of the information read by subsequent frames, thereby effectively preventing them from reproducing the reference image.
    \item We construct two human-centric I2V protection benchmarks with carefully curated reference frames and human-verified motion prompts, and extensively evaluate \method on various I2V models. Results show that \method consistently outperforms existing methods, transfers to models not involved in the optimization, and remains effective against transformations and various attacks.
\end{itemize}

\section{Related Work}
\label{sec:related}

\textbf{Image-to-Video Diffusion Models.} Given a reference image and a prompt, I2V diffusion models aim to generate videos that preserve the appearance of the reference image while following the motion described by the prompt, and they have been widely adopted in content creation and animation~\citep{wang2026image}. Early I2V models, such as SVD~\citep{blattmann2023stable}, are mostly built upon UNet architectures extended from text-to-image models. Constrained by their model capacity and spatiotemporal modeling ability, these models remain limited in generation fidelity and temporal consistency~\citep{wang2026image}. More recent models, such as CogVideoX~\citep{yang2025cogvideox}, Wan~\citep{wan2025wan}, and LTX-Video~\citep{hacohen2024ltx}, shift to DiT architectures with full spatiotemporal attention, substantially improving subject fidelity, generation quality, and temporal coherence. These models typically encode the reference image into latent tokens aligned with the first frame, and subsequent frames primarily interact with these tokens through attention to acquire visual information. This reading pathway plays a key role in video generation, as evidenced by recent long video generation methods~\citep{liu2026rolling,yi2025deep}, which demonstrate that retaining the first-frame tokens as attention anchors can help improve long-range subject consistency. In this work, we show that this reading pathway also offers an efficient lever for protection.

\textbf{Protective Perturbations against Unauthorized Visual Generation.} While generative models have greatly facilitated visual content creation, they are also increasingly misused, e.g., for unauthorized style mimicry and identity forgery~\citep{zhang2026adversarial}. To prevent such misuse, protective perturbations add small adversarial noise to images before release, aiming to block unauthorized generation at the source. In the era of diffusion models, PhotoGuard~\citep{salman2023photoguard} and AdvDM~\citep{liang2023advdm} are among the earliest to protect images from being edited and customized. These methods typically perturb the image's VAE latent or maximize the denoising loss, so that models like Stable Diffusion cannot faithfully process the protected images. To date, this research area remains highly active, with subsequent studies further improving efficiency~\citep{xue2024toward} and enhancing protection in specific scenarios~\citep{li2025styleguard,jia2026adapter}.

With the rapid advancement of I2V models and the growing concerns over their misuse, recent studies have begun to extend protective perturbations to I2V generation~\citep{gui2025i2vguard,song2026antiprompt}. Early efforts mainly target UNet-based models~\citep{pang2024vgmshield,zhou2025dormant}. Since these models largely inherit their components from image diffusion models, these methods naturally build upon designs from image protection, e.g., disrupting the encoders. As DiT-based models become the mainstream, more recent studies further introduce objectives on intermediate features or the denoising process~\citep{long2026immune2v,song2026antiprompt}, demonstrating the promise on modern I2V models. However, designing such objectives is inherently challenging: in DiT-based models, the reference image's features propagate to subsequent frames through deeply stacked spatiotemporal attention, and it remains largely unclear which internal factor governs how much of the reference is reproduced. Without such an understanding, protection objectives have to be defined on proxies that are only indirectly related to this reading process, and part of the budget may be spent on features that contribute little to what subsequent frames read, limiting the defenses' effectiveness. In this paper, we aim to fill this gap by examining how I2V models read the reference image, which allows us to concentrate the budget more directly on the reading process and unlock more effective protection.

\section{Motivating Analysis}
\label{sec:read}

As discussed in Sec.~\ref{sec:related}, understanding how subsequent frames read from the reference image is central towards effective protection. In this section, we first formalize this reading process in modern I2V models and then analyze the factors influencing it, which motivates the design of our protection.

\noindent\textbf{The Reading Pathway.} Given a reference image $x_{\mathrm{ref}}$ and a text prompt $y$, I2V models such as Wan~2.1 and CogVideoX generate a video by starting from Gaussian noise $z_T$ and iteratively denoising it. At each step, the current latent $z_t$ is split into spatio-temporal patches and linearly projected into tokens, on which the network operates. The visual content of $x_{\mathrm{ref}}$ is encoded and carried by $N$ tokens at specific positions (typically those of the first frame),  which we refer to as reference tokens.

Inside the denoising network, the MLP and the normalization layers act on each token separately, and attention is the main operator that moves information between tokens~\citep{yang2025cogvideox,wan2025wan}. Concretely, within one head of one layer, each reference token $i$ is projected into a key vector $k_i = W_K h_i$ and a value vector $v_i = W_V h_i$, where $h_i$ is its hidden state. The key serves for matching and encodes what the token depicts, while the value carries the appearance features that are transferred. In turn, a subsequent-frame token is projected into a query $q_m = W_Q h_m$, which is matched against the keys, and the token receives the values in proportion to the match. For instance, when $x_{\mathrm{ref}}$ depicts a person, the query of the token generating the face in a subsequent frame matches the keys of the facial tokens, and it then receives the facial values, while the token generating the background likewise receives the background values. In this way, each position obtains the content it needs, and the video reproduces the visual features of $x_{\mathrm{ref}}$. Formally, let $o_m$ denote the content that the $m$-th subsequent-frame token reads from the reference tokens in this head (with $\alpha_{mi}$ renormalized within the $N$ reference tokens), we have:
\begin{equation}
\label{eq:readout}
o_m \;=\; \sum_{i=1}^{N} \alpha_{mi}\, v_i, \qquad \alpha_{mi} \;=\; \frac{\exp\!\big(q_m^\top k_i/\sqrt{d}\big)} {\sum_{i'=1}^{N}\exp\!\big(q_m^\top k_{i'}/\sqrt{d}\big)} .
\end{equation}

\noindent\textbf{Theoretical Analysis from a Convex Hull Perspective.} One can observe from Eq.~(\ref{eq:readout}) that the weights $\alpha_{mi}$ are non-negative and sum to one, so that $o_m$ corresponds to a point in the convex hull~\citep{ball1997elementary} of the reference values. Specifically, since $\alpha_{mi} \ge 0$ and $\sum_i \alpha_{mi} = 1$, the weight vector $\alpha_m = (\alpha_{m1},\dots,\alpha_{mN})^\top$ lies in the probability simplex $\Delta^{N-1} := \{\alpha \in \mathbb{R}^{N} \mid \alpha_i \ge 0,\; \sum_i \alpha_i = 1\}$. The map $\alpha \mapsto \sum_i \alpha_i v_i$ is linear and sends this simplex to the convex hull of $\{v_i\}$, with the vertices mapped to the individual values. Therefore, we obtain
\begin{equation}
\label{eq:hull}
o_m \;\in\; \mathcal{H}(V) \;:=\; \mathrm{conv}(v_1,\dots,v_N) \;=\; \Big\{\, \sum\nolimits_{i=1}^{N}\alpha_i v_i \;\Big|\; \alpha \in \Delta^{N-1} \,\Big\},
\end{equation}
i.e., what a subsequent-frame token can read from the reference tokens is confined to the convex hull of the reference tokens' value vectors $\{v_i\}$. Since a convex hull is determined by its vertices, the attention weights can only select a point within it as the readout $o_m$ and cannot read content beyond what $\{v_i\}$ span. In our previous example, the video can reproduce distinct visual content (e.g., the face and the background) because $\mathcal{H}(V)$ is large enough for different queries to reach points that express completely different visual features. Thus, the size of $\mathcal{H}(V)$ is a key quantity that governs the range of content subsequent frames, as formally analyzed in the following proposition:

\begin{proposition}[Value Spread Bounds Readout Diversity]
\label{prop:spread}
Let $\bar{o} = \frac{1}{M}\sum_{m=1}^{M} o_m$, and let $R(V) = \min_{c}\max_{i}\|v_i - c\|$ denote the radius of the minimum enclosing ball of $\{v_1,\dots,v_N\}$. Suppose the $M$ subsequent-frame tokens read from the reference tokens according to Eq.~(\ref{eq:readout}), i.e., $o_m = \sum_{i}\alpha_{mi} v_i$. Then for any $M$ and any attention weights $\alpha_m \in \Delta^{N-1}$,
\begin{equation}
\label{eq:spread}
\frac{1}{M}\sum_{m=1}^{M}\big\|o_m - \bar{o}\big\|^2 \;\le\; R(V)^2 .
\end{equation}
\end{proposition}

Proposition~\ref{prop:spread} shows that the spread of the reference values bounds the variance of their readouts across subsequent-frame tokens. A smaller $R(V)$ imposes a lower ceiling on this diversity, regardless of how attention is distributed over the reference tokens. In the extreme case where all reference values coincide, every normalized reference readout in this head equals the same vector. This observation motivates reducing reference-value diversity to constrain the reading pathway.

\noindent\textbf{Empirical Validation.} Proposition~\ref{prop:spread} shows that, in a single attention operation, the diversity of the reference values bounds what subsequent-frame tokens can read from them. However, in reality, the generation process involves multiple layers, denoising steps, and residual connections, and how the quality and fidelity of the generated video depend on the diversity of the reference values remains unclear from the bound alone. To empirically validate its impact, we conduct an intervention experiment on real video generation models, in which we directly contract the values of the reference tokens and observe how the generated video changes. Specifically, we replace the value of each reference token with $v_i \leftarrow c + \gamma\,(v_i - c)$, where $c$ is the center of the minimum enclosing ball and $\gamma \in [0, 1]$, so that $R(V)$ shrinks linearly with $\gamma$, with $\gamma = 1$ recovering the original model and $\gamma = 0$ corresponding to the extreme case where the hull degenerates to a point. The intervention is applied at every layer and the forward pass otherwise proceeds unchanged, with no other component (queries, keys, text condition, or noise seed) intervened.

\begin{figure}[t]
\centering
\includegraphics[width=0.98\linewidth]{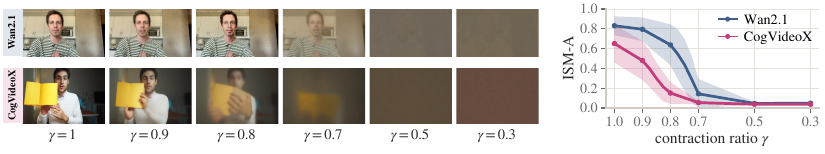}
\vspace{-1.2em}
\caption{Effect of contracting the reference values toward their center by a ratio $\gamma$. \textbf{Left}: generated frames under different $\gamma$. \textbf{Right}: the ArcFace identity similarity (ISM-A) between generated frames and the reference, versus $\gamma$. A mild contraction already severely degrades the generated videos.}\label{fig:intervene}
\vspace{-1em}
\end{figure}

As shown in Fig.~\ref{fig:intervene}, contracting $R(V)$ by only $20\%$ ($\gamma = 0.8$) already introduces visible artifacts on the face, and at $30\%$ ($\gamma = 0.7$) the video largely collapses into degenerated color blocks with no recognizable subject. Moreover, even when the intervention is applied to only a subset of layers and denoising steps, a contraction of $30\%$ also suffices to almost completely destroy the video. We also verify that random perturbations of the same magnitude do not cause comparable degradation. These results confirm that reducing the diversity of the reference values is an efficient lever for protection.

\section{Methodology}
\label{sec:method}

\noindent\textbf{Problem Formulation.} We formulate the problem as follows. An image owner (the defender) wants to publish an image $x_{\mathrm{ref}}$ (e.g., a selfie), and does not want it to be used as the reference image of an I2V model to fabricate videos. To this end, the defender releases a protected image $x' = x_{\mathrm{ref}} + \delta$. Following prior work on protective perturbations~\citep{gui2025i2vguard,song2026antiprompt}, we adopt the commonly used $\ell_\infty$ budget with $\|\delta\|_\infty \le \epsilon$. Then, an attacker may collect $x'$ from the Internet and feed it, possibly after input transformations, into an I2V model with an arbitrary prompt $y$ and random noise $z_T$. The defender's goal is to craft the protective perturbation $\delta$ such that the generated video can no longer faithfully reproduce $x_{\mathrm{ref}}$ with high quality.

\noindent\textbf{Value Collapse Objective.}
Proposition~\ref{prop:spread} identifies the spread of the reference values as an upper bound on reference-readout diversity, while our intervention experiment shows that contracting these values can substantially disrupt generation. This motivates us to disrupt video generation by collapsing the values of the reference tokens. However, $R(V)$ is inconvenient to optimize directly: computing it requires solving a minimum enclosing ball problem at every forward pass, and its gradient is sparse, being nonzero only for the few values lying on the boundary. Moreover, the absolute scale of $R(V)$ varies across layers and heads, which makes it hard to be aggregated.

To this end, we propose a value collapse loss, which retains the target of $R(V)$ while avoiding the aforementioned difficulties. Let $V^{(\ell,h,t)}(x') = \{v_i\}_{i=1}^{N}$ denote the values of the reference tokens at layer $\ell$, head $h$, and denoising step $t$ of the surrogate model, $\bar{v}$ their mean, and $\mathcal{B}$ the set of layers and heads on which the objectives are imposed. We define the value collapse loss $\mathcal{L}_{\mathrm{vc}}$ as follows:
\begin{equation}
\label{eq:vc}
\mathcal{L}_{\mathrm{vc}}(\delta) \;=\; \mathbb{E}_{t}\sum_{(\ell,h)\in\mathcal{B}}
\frac{\sum_{i=1}^{N}\big\|v_i^{(\ell,h,t)}(x_{\mathrm{ref}}+\delta)-\bar{v}^{(\ell,h,t)}\big\|^2}
{\sum_{i=1}^{N}\big\|v_i^{(\ell,h,t)}(x_{\mathrm{ref}}+\delta)\big\|^2}.
\end{equation}

Intuitively, since $\sum_i\|v_i\|^2 = \sum_i\|v_i-\bar{v}\|^2 + N\|\bar{v}\|^2$, each term in the sum is the fraction of the total value energy that is not shared by all reference tokens. Minimizing it pushes the energy onto the common component $\bar{v}$ and drives all reference tokens toward a single vector. Unlike $R(V)$, whose gradient involves only the few most distant values, this objective has a gradient involving all reference tokens. Moreover, for nonzero total value energy, each term is differentiable, bounded in $[0,1]$, and invariant to a common nonzero scaling of the values, which allows direct aggregation across layers and heads without additional per-layer calibration.

\noindent\textbf{Attention Attraction Objective.}
In our previous analysis, the attention weights in Eq.~(\ref{eq:readout}) are normalized within the reference tokens. However, in practice, a subsequent-frame token distributes its attention over the entire sequence, including other video tokens and embedding tokens of the reference image~\citep{wang2026image}. Thus, how much the collapsed reference values contribute to its output depends on the attention allocated to them, which is not directly controlled by $\mathcal{L}_{\mathrm{vc}}$.

To further amplify and stabilize the collapse's effect, we introduce another objective that attracts the attention of subsequent-frame tokens to the reference tokens. Our key insight is that the attention mass is governed by the logits between the subsequent-frame queries and the keys of the reference tokens, which could be influenced through $\delta$. Thus, we propose to raise the average attention logit from the subsequent-frame queries to the reference keys. Let $\bar{q}^{(\ell,h,t)} = \frac{1}{M}\sum_{m} q_m$ denote the mean query of the subsequent-frame tokens and $\bar{k}^{(\ell,h,t)}(x_{\mathrm{ref}}+\delta) = \frac{1}{N}\sum_{i} k_i$ the mean key of the reference tokens at layer $\ell$, head $h$, and denoising step $t$. The attention attraction loss $\mathcal{L}_{\mathrm{aa}}$ is defined as:
\begin{equation}
\label{eq:aa}
\mathcal{L}_{\mathrm{aa}}(\delta) \;=\; -\,\mathbb{E}_{t}\sum_{(\ell,h)\in\mathcal{B}}
\frac{\bar{q}^{(\ell,h,t)\top}\,\bar{k}^{(\ell,h,t)}(x_{\mathrm{ref}}+\delta)}{\sqrt{d}}.
\end{equation}
Intuitively, $\bar{q}^\top\bar{k}/\sqrt{d}$ is the average attention logit from the subsequent-frame queries to the reference keys, and raising it tends to increase the share of attention received by the reference tokens~\citep{kim2023dense}. We treat $\bar{q}$ as a constant, since the queries are given by the attacker's generation process and the perturbation should act through the keys alone, which also allows $\mathcal{L}_{\mathrm{aa}}$ to be efficiently computed. Moreover, since the keys are usually QK-normalized~\citep{wan2025wan,yang2025cogvideox}, the average logit is bounded, and $\mathcal{L}_{\mathrm{aa}}$ cannot be reduced by merely inflating the key norm.

\textbf{Overall Objective.} To get the final protected image, we obtain the perturbation $\delta$ by jointly optimizing the two objectives:
\begin{equation}
\label{eq:overall}
\min_{\|\delta\|_\infty \le \epsilon}\ \mathcal{L}_{\mathrm{Lethe}}(\delta) \;=\; \mathcal{L}_{\mathrm{vc}}(\delta) + \alpha\,\mathcal{L}_{\mathrm{aa}}(\delta),
\end{equation}
where $\epsilon$ is the perturbation budget and $\alpha$ balances the two terms. Intuitively, $\mathcal{L}_{\mathrm{vc}}$ and $\mathcal{L}_{\mathrm{aa}}$ form a complementary pair that jointly shapes the reference-frame tokens. Specifically, $\mathcal{L}_{\mathrm{vc}}$ suppresses deviations of the reference values from their mean, making them provide subsequent frames with highly homogeneous values. Complementarily, $\mathcal{L}_{\mathrm{aa}}$ steers the reference keys, so that the readout of subsequent frames is largely drawn from these impoverished tokens. Together, \method leaves subsequent frames with little distinguishable information from the reference image and limited capacity to compensate through other tokens, leading to notable and persistent degradation in the generated video. In practice, we solve Eq.~(\ref{eq:overall}) with MI-FGSM~\citep{dong2018boosting} for 100 iterations.

\section{Experiments}
\label{sec:exp}

\subsection{Experimental Setup}
\label{sec:impl}

\noindent\textbf{Models.} We mainly conduct our experiments on 2 SOTA native image-to-video generative models, i.e., Wan2.1-I2V-14B~\citep{wan2025wan} and CogVideoX-5B-I2V~\citep{yang2025cogvideox}, which represent the mainstream DiT-based paradigm with distinct training objectives and attention designs. To evaluate transferability, we also evaluate on LTX-Video~\citep{hacohen2024ltx}, SVD~\citep{blattmann2023stable}, LightX2V~\citep{lightx2v}, and AccVideo~\citep{zhang2025accvideo}, covering models with completely different architectures and training data, as well as popular community models.

\noindent\textbf{Baselines.} We compare \method with 5 representative baselines that protect images against diffusion models, including PhotoGuard~\citep{salman2023photoguard}, AdvDM~\citep{liang2023advdm}, SDS+~\citep{xue2024toward}, I2VGuard~\citep{gui2025i2vguard}, and Anti-Prompt~\citep{song2026antiprompt}, where Anti-Prompt is a very recent work published in 2026 and represents the state of the art that specifically targets disrupting I2V generation. All baselines are reproduced following their original papers and official code (if open sourced) with recommended hyperparameters. To ensure fairness, all methods share the same perturbation budget ($\epsilon = 16/255$ by default) and inference configurations.

\noindent\textbf{Datasets.} Existing I2V benchmarks~\citep{huang2024vbench} mainly target general-purpose content with few samples of human faces and bodies, while human-centric datasets~\citep{yu2023celebv,carreira2019kinetics} carry annotations designed for other tasks (e.g., attribute descriptions or action labels) that cannot directly serve as I2V prompts. Therefore, we construct and annotate two evaluation datasets built upon CelebV-Text~\citep{yu2023celebv} and Kinetics-700~\citep{carreira2019kinetics}, which correspond to close-up facial portraits and human actions in natural scenes, respectively. For each dataset, we select 350 high-quality videos with distinct identities, extract one frame with a clearly visible face from each video as the conditioning image, and annotate each image with several I2V prompts describing the appearance and the motion of the subject. We will release the re-annotated datasets to facilitate future research on human-centric I2V protection.

\noindent\textbf{Metrics.} We evaluate protection performance from both identity preservation and generation quality perspectives. For identity preservation, we report the Identity Score Matching (ISM, $\downarrow$)~\citep{van2023antidb,huang2026delving}, i.e., the cosine similarity between the face embeddings of the generated frames and that of the clean conditioning image, averaged over frames. To reduce the dependence on a single recognizer, we report ISM under two face recognition models, ArcFace~\citep{deng2019arcface} and FaceNet~\citep{schroff2015facenet}, denoted as ISM-A and ISM-F, respectively. For generation quality, we adopt Q-Align~\citep{wu2024qalign} to score the generated videos at both the frame level (QAF, $\downarrow$), which reflects the visual quality of individual frames, and the video level (QAV, $\downarrow$), which further accounts for video-level quality (e.g., temporal consistency and motion naturalness). Lower ISM, QAF, and QAV indicate that the generated video preserves less identity and has lower visual quality, and thus a stronger protection. Additionally, following \cite{song2026antiprompt}, we adopt an MLLM-based evaluation protocol as a complement to the above metrics.

\begin{table*}[t]
\centering
\small
\setlength{\tabcolsep}{3.4pt}
\caption{\textbf{Main Results.} White-box protection performance on Wan2.1 and CogVideoX on the CelebV-Text and Kinetics-700 datasets. All methods use an \(\ell_\infty\) budget of \(\epsilon=16/255\).}
\label{tab:main}
\begin{tabular}{c l cccc cccc}
\toprule
& & \multicolumn{4}{c}{\textbf{CelebV-Text}} & \multicolumn{4}{c}{\textbf{Kinetics-700}} \\
\cmidrule(lr){3-6} \cmidrule(lr){7-10}
Model & Method & ISM-A \(\downarrow\) & ISM-F \(\downarrow\) & QAF \(\downarrow\) & QAV \(\downarrow\) & ISM-A \(\downarrow\) & ISM-F \(\downarrow\) & QAF \(\downarrow\) & QAV \(\downarrow\) \\
\midrule
\multirow{7}{*}{\rotatebox[origin=c]{90}{Wan2.1}}
& Clean       & 0.8173 & 0.8725 & 0.9123 & 0.9462 & 0.7584 & 0.8242 & 0.7974 & 0.8381 \\
& AdvDM       & 0.6082 & 0.7470 & 0.6199 & 0.6866 & 0.5621 & 0.6914 & 0.5623 & 0.6176 \\
& SDS+        & 0.6758 & 0.8018 & 0.5894 & 0.6469 & 0.6119 & 0.7377 & 0.5449 & 0.5969 \\
& PhotoGuard  & 0.7344 & 0.8224 & 0.6368 & 0.6967 & 0.6945 & 0.7747 & 0.5824 & 0.6404 \\
& I2VGuard    & 0.5818 & 0.7263 & 0.6285 & 0.7045 & 0.5666 & 0.6952 & 0.5843 & 0.6497 \\
& Anti-Prompt  & 0.5237 & 0.7122 & 0.8447 & 0.8985 & 0.4956 & 0.6663 & 0.7341 & 0.8011 \\
\rowcolor{gray!20}
& \textbf{Ours} & \textbf{0.4280} & \textbf{0.5308} & \textbf{0.3970} & \textbf{0.4976} & \textbf{0.4125} & \textbf{0.4957} & \textbf{0.3852} & \textbf{0.4745} \\
\midrule
\multirow{7}{*}{\rotatebox[origin=c]{90}{CogVideoX}}
& Clean       & 0.6611 & 0.7785 & 0.8561 & 0.9123 & 0.5742 & 0.7105 & 0.7623 & 0.8198 \\
& AdvDM       & 0.4356 & 0.6339 & 0.5315 & 0.5750 & 0.3854 & 0.5762 & 0.5191 & 0.5514 \\
& SDS+        & 0.4093 & 0.6104 & 0.5050 & 0.5549 & 0.3729 & 0.5617 & 0.4879 & 0.5273 \\
& PhotoGuard  & 0.4440 & 0.6236 & 0.5538 & 0.6532 & 0.4302 & 0.5912 & 0.5619 & 0.6519 \\
& I2VGuard    & 0.4920 & 0.6831 & 0.7035 & 0.7751 & 0.4535 & 0.6215 & 0.5763 & 0.6527 \\
& Anti-Prompt  & 0.3424 & 0.5598 & 0.7131 & 0.7769 & 0.3522 & 0.5621 & 0.7046 & 0.7650 \\
\rowcolor{gray!20}
& \textbf{Ours} & \textbf{0.2968} & \textbf{0.4895} & \textbf{0.4235} & \textbf{0.4946} & \textbf{0.2523} & \textbf{0.4299} & \textbf{0.4223} & \textbf{0.4828 } \\
\bottomrule
\end{tabular}
\end{table*}

\begin{figure}
    \centering
    \vspace{-0.5em}
    \includegraphics[width=1\linewidth]{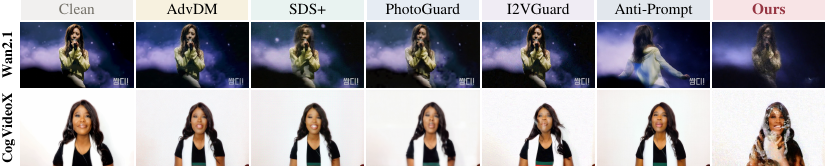}
    \vspace{-1.8em}
    \caption{\textbf{Qualitative comparison} between \method and baselines on Wan2.1 and CogVideoX.}
    \vspace{-1.5em}
    \label{fig:qualitative}
\end{figure}

\subsection{Overall Performance}
\label{sec:main}

\noindent\textbf{Main Results.} We first evaluate all methods on Wan2.1-I2V-14B~\citep{wan2025wan} and CogVideoX-5B-I2V~\citep{yang2025cogvideox} in the white-box setting. As shown in Tab.~\ref{tab:main}, \method achieves the best performance across all 16 settings spanning 2 models, 2 datasets, and 4 metrics, outperforming all evaluated baselines by a notable margin. For instance, on CelebV-Text with Wan2.1, \method reduces ISM-A from 0.817 on clean images to 0.428 and QAF from 0.912 to 0.397, well below the second-best baselines (0.524 and 0.589, respectively). The qualitative results in Fig.~\ref{fig:qualitative} further show that \method degrades both the identity and the visual structure of the generated videos more completely, where it erases the subject on Wan2.1 and smears it into unrecognizable textures on CogVideoX. These results demonstrate that our value collapse and attention attraction strategies effectively block subsequent frames from reading the identity and appearance of the reference image, so that protected images can no longer be faithfully animated into high-quality videos by unauthorized users.

\begin{figure}[t]
\centering
\includegraphics[width=\linewidth]{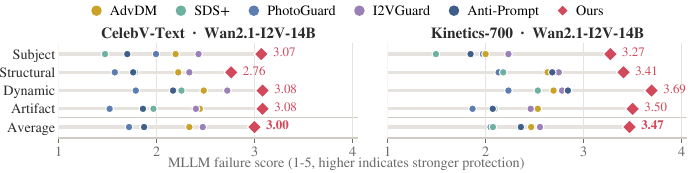}
\vspace{-2em}
\caption{\textbf{MLLM-based evaluation} on Wan2.1.}
\label{fig:vlmjudge}
\end{figure}

\noindent\textbf{MLLM-based Evaluation.} For a more comprehensive evaluation, we adopt the failure-oriented MLLM-based protocol of Anti-Prompt~\citep{song2026antiprompt}, where Qwen3-VL-8B-Instruct~\citep{bai2025qwen3} serves as the judge. The judge is given the generated video, the corresponding prompt, and the original input image, and scores four aspects: \emph{Subject Preservation} (whether the subject remains the recognizable person the same as the one in the input image), \emph{Structural Consistency} (whether shapes and proportions are stable), \emph{Dynamic Consistency} (whether the motion is temporally coherent), and \emph{Artifact Suppression} (whether rendering artifacts are present). Scores range from 1 to 5, and we invert them so that higher values indicate stronger protection. As shown in Fig.~\ref{fig:vlmjudge}, \method achieves the highest score over the four aspects on both datasets. This further confirms that \method provides stronger and more comprehensive protection against unauthorized I2V generation.

\begin{figure}[t]
\centering
\includegraphics[width=\linewidth]{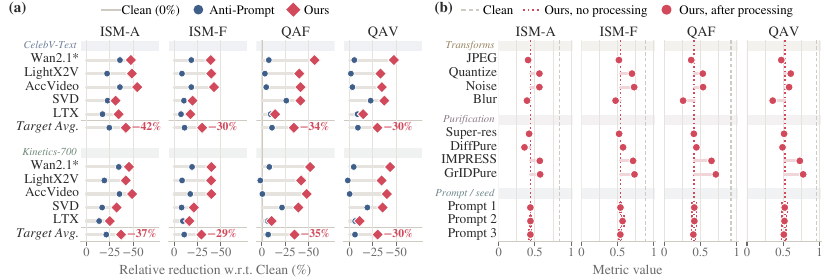}
\vspace{-2em}
\caption{\textbf{Transferability and robustness results.} (a) Relative reduction of each metric w.r.t. clean images when the perturbations crafted on Wan2.1 are applied to four other I2V models. (b) Metric values on Wan2.1 (CelebV-Text) under different input transformations and generation conditions.}

\label{fig:transfer}
\vspace{-1.5em}
\end{figure}

\subsection{Transferability \& Robustness}
\label{sec:rob}

\noindent\textbf{Transferability.} In practice, an attacker may not use exactly the same model as the surrogate. To account for this, we further evaluate whether the perturbations crafted on Wan2.1 transfer to models that are not directly involved in the optimization, and compare \method with Anti-Prompt. Fig.~\ref{fig:transfer}(a) reports the relative reduction of each metric with respect to the clean images. On LightX2V and AccVideo, which are built on the same DiT architecture as Wan2.1, \method retains most of its effectiveness (e.g., lowering all metrics by $\sim40\%$). On SVD and LTX-Video, which differ substantially from Wan2.1 in architecture, \method still yields clear degradation (e.g., $\sim30\%$ ISM reduction for both models) and consistently outperforms Anti-Prompt on all four metrics. We \emph{hypothesize} that this is because tokens carrying reference information are also aggregated through softmax attention in all these models, so the reduced diversity of the reference representation induced by \method also generalizes beyond the surrogate. We leave a more thorough investigation to future work.

\noindent\textbf{Robustness.} Before being fed into the I2V model, a protected image may undergo image transformations (e.g., enhancement by platform), deliberate adversarial purification by the attacker, and generation with prompts and seeds unknown to the defender. To test \method's robustness under these conditions, following \cite{vu2026antii2v}, we evaluate three groups of settings on Wan2.1: (I) 4 common input transformations, i.e., JPEG compression ($q=70$), 3-bit color quantization, Gaussian noise ($\sigma=0.2$), and Gaussian blur (kernel size 7); (II) 4 representative purifiers, i.e., Super-Res~\citep{mustafa2019image}, DiffPure~\citep{diffpure}, IMPRESS~\citep{impress}, and GrIDPure~\citep{GrIDPure}, each with its recommended configuration; and (III) 3 different prompts, each generated with 3 random seeds. As shown in Fig.~\ref{fig:transfer}(b), \method remains highly effective under these variations, and even the purifiers specifically designed to remove protective perturbations (IMPRESS and GrIDPure) recover only part of the gap to clean images, e.g., ISM-A rises from $0.43$ to $\sim0.56$, which is still far below the clean ones ($\sim0.80$). Across prompts and seeds, the metrics also vary by no more than $\sim0.03$, suggesting that \method's effectiveness does not rely on specific generation conditions.

\begin{figure}[t]
\centering
\begin{minipage}[t]{0.535\linewidth}
\vspace{3pt}
\centering
\setlength{\abovecaptionskip}{0pt}
\setlength{\belowcaptionskip}{6pt}
\captionof{table}{\textbf{Ablation on the two objectives of \method} on Wan2.1. Each value is the mean of three runs with different random seeds. Best results are in bold.}
\label{tab:ablation}
\footnotesize
\setlength{\tabcolsep}{3pt}
\begin{tabular}{@{}l cccc@{}}
\toprule
Variant & ISM-A $\downarrow$ & ISM-F $\downarrow$ & QAF $\downarrow$ & QAV $\downarrow$ \\
\midrule
\multicolumn{5}{@{}l}{\textit{CelebV-Text}} \\
Clean & 0.8173 & 0.8725 & 0.9123 & 0.9462 \\
$\mathcal{L}_{\mathrm{vc}}$ only & 0.5899 & 0.6827 & 0.5224 & 0.6009 \\
\rowcolor{gray!12}
$\mathcal{L}_{\mathrm{vc}}+\mathcal{L}_{\mathrm{aa}}$ (\method) & \textbf{0.4280} & \textbf{0.5308} & \textbf{0.3970} & \textbf{0.4976} \\
\midrule
\multicolumn{5}{@{}l}{\textit{Kinetics-700}} \\
Clean & 0.7584 & 0.8242 & 0.7974 & 0.8381 \\
$\mathcal{L}_{\mathrm{vc}}$ only & 0.5084 & 0.6127  & 0.4600 & 0.5324 \\
\rowcolor{gray!12}
$\mathcal{L}_{\mathrm{vc}}+\mathcal{L}_{\mathrm{aa}}$ (\method) & \textbf{0.4125} & \textbf{0.4957} & \textbf{0.3852} & \textbf{0.4745} \\
\bottomrule
\end{tabular}
\end{minipage}\hfill
\begin{minipage}[t]{0.44\linewidth}
\vspace{0pt}
\centering
\setlength{\abovecaptionskip}{2pt}
\includegraphics[width=\linewidth]{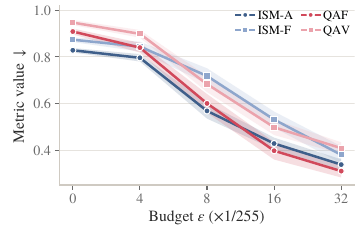}
\caption{\textbf{Effect of the budget} $\epsilon$ on Wan2.1 and CelebV-Text. Shaded bands denote one standard deviation over three random seeds.}
\label{fig:eps}
\end{minipage}
\vspace{-1.0em}
\end{figure}

\subsection{Ablation Studies}
\label{sec:ablation}
\noindent\textbf{Effect of each objective.} We first ablate the two objectives. As shown in Tab.~\ref{tab:ablation}, using $\mathcal{L}_{\mathrm{vc}}$ alone already provides strong protection, e.g., reducing ISM-A from 0.817 to 0.590 and QAF from 0.912 to 0.522 on CelebV-Text, respectively. Adding $\mathcal{L}_{\mathrm{aa}}$ further amplifies the protection by a notable margin, as reflected by consistently lower values on all four metrics across both datasets. This is consistent with our analysis in Sec.~\ref{sec:method} that $\mathcal{L}_{\mathrm{vc}}$ primarily collapses the convex hull, whereas $\mathcal{L}_{\mathrm{aa}}$ keeps the attention on these  tokens, which amplifies the collapse and makes the protection stronger.

\noindent\textbf{Effect of the budget.} Fig.~\ref{fig:eps} shows how each metric varies with the perturbation budget $\epsilon$. Consistent with the common observation in this field~\citep{liang2023advdm,long2026immune2v}, a larger budget yields stronger protection, and all four metrics decrease monotonically with $\epsilon$. Notably, with only half of the perturbation budget ($\epsilon = 8/255$), \method already achieves protection comparable to several strong baselines at $\epsilon = 16/255$, e.g., an ISM-A of 0.58 and a QAV of 0.69. This suggests that \method spends the perturbation budget more efficiently, as both of its objectives directly target the pathway through which subsequent frames read the reference image.

\section{Conclusion}
This paper studies how to protect personal images against unauthorized I2V generation. We first show that subsequent frames read visual content from the reference frame primarily with attention, and demonstrate through a convex hull-based view that the diversity of these readouts is bounded by the spread of the reference values. Our intervention experiment further confirms that contracting this spread effectively disrupts generation. Building on this analysis, we propose \method, which consists of a value collapse loss that suppresses the diversity of the reference values and an attention attraction loss that draws the attention of subsequent frames to the collapsed reference tokens, so that the protected images can no longer be faithfully animated. Extensive experiments on multiple I2V models and two re-annotated human-centric datasets demonstrate \method's strong performance in various settings. We hope \method, together with our analyses and the dataset can deepen the community's understanding and inspire more effective image protection methods in the future.%

\subsection*{Ethics statement}
This work is motivated by the growing concern that personal images can be used without authorization to generate videos of the depicted person through image-to-video models. By analyzing how these models read visual content from the reference image, we hope to provide insights into what governs the effectiveness of protective perturbations in this setting, and we believe \method represents a valuable step toward more effective protection of personal images. All models used in this work are publicly released and used in accordance with their licenses. Our evaluation sets are built upon CelebV-Text and Kinetics-700, which we use for non-commercial research in accordance with their terms of use. Our re-annotation only adds textual descriptions and collects no new personal data. To comply with the original terms, we will release only the textual annotations together with the video identifiers and frame indices, so that users obtain the original data from official sources. The qualitative examples in this paper are limited to those needed to illustrate the protection effect.

\bibliography{iclr2027_conference}

@article{wan2025wan,
  title={Wan: Open and advanced large-scale video generative models},
  author={Wan, Team and Wang, Ang and Ai, Baole and Wen, Bin and Mao, Chaojie and Xie, Chen-Wei and Chen, Di and Yu, Feiwu and Zhao, Haiming and Yang, Jianxiao and others},
  journal={arXiv preprint arXiv:2503.20314},
  year={2025}
}

@inproceedings{yang2025cogvideox,
  title={Cogvideox: Text-to-video diffusion models with an expert transformer},
  author={Yang, Zhuoyi and Teng, Jiayan and Zheng, Wendi and Ding, Ming and Huang, Shiyu and Xu, Jiazheng and Yang, Yuanming and Hong, Wenyi and Zhang, Xiaohan and Feng, Guanyu and others},
  booktitle={International Conference on Learning Representations},
  year={2025}
}

@article{hacohen2024ltx,
  title={Ltx-video: Realtime video latent diffusion},
  author={HaCohen, Yoav and Chiprut, Nisan and Brazowski, Benny and Shalem, Daniel and Moshe, Dudu and Richardson, Eitan and Levin, Eran and Shiran, Guy and Zabari, Nir and Gordon, Ori and others},
  journal={arXiv preprint arXiv:2501.00103},
  year={2024}
}

@article{zhang2025accvideo,
  title={Accvideo: Accelerating video diffusion model with synthetic dataset},
  author={Zhang, Haiyu and Chen, Xinyuan and Wang, Yaohui and Liu, Xihui and Wang, Yunhong and Qiao, Yu},
  journal={arXiv preprint arXiv:2503.19462},
  year={2025}
}

@article{blattmann2023stable,
  title={Stable video diffusion: Scaling latent video diffusion models to large datasets},
  author={Blattmann, Andreas and Dockhorn, Tim and Kulal, Sumith and Mendelevitch, Daniel and Kilian, Maciej and Lorenz, Dominik and Levi, Yam and English, Zion and Voleti, Vikram and Letts, Adam and others},
  journal={arXiv preprint arXiv:2311.15127},
  year={2023}
}

@misc{lightx2v,
  author       = {{LightX2V Team}},
  title        = {{LightX2V}: {AI} Video Generation and Multimodal Content Production Platform},
  year         = {2026},
  publisher    = {Light AI},
  howpublished = {\url{https://x2v.light-ai.top/}},
  note         = {Accessed: 2026-09-22},
}

@inproceedings{song2026antiprompt,
  title     = {Anti-Prompt: Image Protection against Text-Guided Image-to-Video Generation},
  author    = {Song, Yeonghwan and Lee, Chanhui and Park, Jinsoo and Son, Jeany},
  booktitle = {European Conference on Computer Vision (ECCV)},
  year      = {2026}
}

@InProceedings{diffpure,
  title = 	 {Diffusion Models for Adversarial Purification},
  author =       {Nie, Weili and Guo, Brandon and Huang, Yujia and Xiao, Chaowei and Vahdat, Arash and Anandkumar, Animashree},
  booktitle = 	 {Proceedings of the 39th International Conference on Machine Learning},
  pages = 	 {16805--16827},
  year = 	 {2022},
  editor = 	 {Chaudhuri, Kamalika and Jegelka, Stefanie and Song, Le and Szepesvari, Csaba and Niu, Gang and Sabato, Sivan},
  volume = 	 {162},
  series = 	 {Proceedings of Machine Learning Research},
  month = 	 {17--23 Jul},
  publisher =    {PMLR},
  url = 	 {https://proceedings.mlr.press/v162/nie22a.html}
}

@article{mustafa2019image,
  title={Image super-resolution as a defense against adversarial attacks},
  author={Mustafa, Aamir and Khan, Salman H and Hayat, Munawar and Shen, Jianbing and Shao, Ling},
  journal={IEEE Transactions on Image Processing},
  volume={29},
  pages={1711--1724},
  year={2019},
  publisher={IEEE}
}

@article{pang2024vgmshield,
  title={Vgmshield: Mitigating misuse of video generative models},
  author={Pang, Yan and Chen, Baicheng and Zhang, Yang and Wang, Tianhao},
  journal={arXiv preprint arXiv:2402.13126},
  year={2024}
}

@article{huang2026delving,
  title={Delving into the Temporal Challenges of Unified Video Protection Against Image-to-Video and Fine-Tuning-based Customization},
  author={Huang, Yuxin and Hong, Ziming and Gong, Mingming and Wang, Wanyu and Zhang, Jing and Liu, Tongliang},
  journal={arXiv preprint arXiv:2607.13336},
  year={2026}
}

@inproceedings{jia2026adapter,
  title={Adapter Shield: A Unified Framework with Built-in Authentication for Preventing Unauthorized Zero-Shot Image-to-Image Generation},
  author={Jia, Jun and Miao, Hongyi and Zhou, Yingjie and Zhou, Wangqiu and Zhang, Jianbo and Cao, Linhan and Zhu, Dandan and Yang, Hua and Min, Xiongkuo and Sun, Wei and others},
  booktitle={Proceedings of the IEEE/CVF Conference on Computer Vision and Pattern Recognition},
  pages={30120--30129},
  year={2026}
}

@inproceedings{li2025styleguard,
  title={StyleGuard: Preventing Text-to-Image-Model-based Style Mimicry Attacks by Style Perturbations},
  author={Li, Yanjie and Zhang, Wenxuan and Lyu, Xinqi and Liu, Yihao and Xiao, Bin},
  booktitle={The Thirty-ninth Annual Conference on Neural Information Processing Systems},
  year={2025}
}

@article{wang2026image,
  title={Image-to-Video Diffusion: From Foundations to Open Frontiers},
  author={Wang, Xianlong and Pan, Wenbo and Zhou, Shijia and Li, Ke and Wang, Yuqi and Ye, Zeyu and Zhang, Hangtao and Zhang, Leo Yu and Jia, Xiaohua},
  journal={arXiv preprint arXiv:2605.17248},
  year={2026}
}

@article{yi2025deep,
  title={Deep forcing: Training-free long video generation with deep sink and participative compression},
  author={Yi, Jung and Jang, Wooseok and Cho, Paul Hyunbin and Nam, Jisu and Yoon, Heeji and Kim, Seungryong},
  journal={arXiv preprint arXiv:2512.05081},
  year={2025}
}

@inproceedings{liu2026rolling,
  title={Rolling forcing: Autoregressive long video diffusion in real time},
  author={Liu, Kunhao and Hu, Wenbo and Xu, Jiale and Shan, Ying and Lu, Shijian},
  booktitle={International Conference on Learning Representations},
  volume={2026},
  pages={91177--91196},
  year={2026}
}

@article{ball1997elementary,
  title={An elementary introduction to modern convex geometry},
  author={Ball, Keith and others},
  journal={Flavors of geometry},
  volume={31},
  number={1-58},
  pages={26},
  year={1997}
}

@inproceedings{deng2019arcface,
  title={Arcface: Additive angular margin loss for deep face recognition},
  author={Deng, Jiankang and Guo, Jia and Xue, Niannan and Zafeiriou, Stefanos},
  booktitle={Proceedings of the IEEE/CVF conference on computer vision and pattern recognition},
  pages={4690--4699},
  year={2019}
}

@article{shao2026efficient,
  title={Efficient video diffusion models: Advancements and challenges},
  author={Shao, Shitong and Bai, Lichen and Wan, Pengfei and Kwok, James and Xie, Zeke},
  journal={arXiv preprint arXiv:2604.15911},
  year={2026}
}

@article{zhang2026adversarial,
  title={Adversarial Attacks for Good: A Survey of Proactive Protection across the Visual Content Lifecycle},
  author={Zhang, Jiaming and Chen, Boyang and Li, Zherui and Zhang, Fuyao and Yan, Xinyu and Tae, Hong Xi and He, Wenwen and Wang, Xuan and Guo, Siqi and Dong, Junhao and others},
  journal={arXiv preprint arXiv:2608.04314},
  year={2026}
}

@inproceedings{zhou2025dormant,
  title={Dormant: Defending against pose-driven human image animation},
  author={Zhou, Jiachen and Wang, Mingsi and Li, Tianlin and Meng, Guozhu and Chen, Kai},
  booktitle={34th USENIX Security Symposium (USENIX Security 25)},
  pages={5209--5228},
  year={2025}
}

@article{szegedy2013intriguing,
  title={Intriguing properties of neural networks},
  author={Szegedy, Christian and Zaremba, Wojciech and Sutskever, Ilya and Bruna, Joan and Erhan, Dumitru and Goodfellow, Ian and Fergus, Rob},
  journal={arXiv preprint arXiv:1312.6199},
  year={2013}
}

@article{kamachee2025video,
  title={Video deepfake abuse: How company choices predictably shape misuse patterns},
  author={Kamachee, Max and Casper, Stephen and Ding, Michelle L and Yew, Rui-Jie and Reuel, Anka and Biderman, Stella and Hadfield-Menell, Dylan},
  journal={arXiv preprint arXiv:2512.11815},
  year={2025}
}

@article{long2026immune2v,
  title={Immune2V: Image Immunization Against Dual-Stream Image-to-Video Generation},
  author={Long, Zeqian and Kara, Ozgur and Xue, Haotian and Chen, Yongxin and Rehg, James M},
  journal={arXiv preprint arXiv:2604.10837},
  year={2026}
}

@inproceedings{gridpure,
  title={Can protective perturbation safeguard personal data from being exploited by stable diffusion?},
  author={Zhao, Zhengyue and Duan, Jinhao and Xu, Kaidi and Wang, Chenan and Zhang, Rui and Du, Zidong and Guo, Qi and Hu, Xing},
  booktitle={2024 IEEE/CVF Conference on Computer Vision and Pattern Recognition (CVPR)},
  pages={24398--24407},
  year={2024},
  organization={IEEE}
}

@article{impress,
  title={Impress: Evaluating the resilience of imperceptible perturbations against unauthorized data usage in diffusion-based generative ai},
  author={Cao, Bochuan and Li, Changjiang and Wang, Ting and Jia, Jinyuan and Li, Bo and Chen, Jinghui},
  journal={Advances in Neural Information Processing Systems},
  volume={36},
  pages={10657--10677},
  year={2023}
}

@inproceedings{gui2025i2vguard,
  title={I2vguard: Safeguarding images against misuse in diffusion-based image-to-video models},
  author={Gui, Dongnan and Guo, Xun and Zhou, Wengang and Lu, Yan},
  booktitle={2025 IEEE/CVF Conference on Computer Vision and Pattern Recognition (CVPR)},
  pages={12595--12604},
  year={2025},
  organization={IEEE}
}

@InProceedings{liang2023advdm,
  title = 	 {Adversarial Example Does Good: Preventing Painting Imitation from Diffusion Models via Adversarial Examples},
  author =       {Liang, Chumeng and Wu, Xiaoyu and Hua, Yang and Zhang, Jiaru and Xue, Yiming and Song, Tao and Xue, Zhengui and Ma, Ruhui and Guan, Haibing},
  booktitle = 	 {Proceedings of the 40th International Conference on Machine Learning},
  pages = 	 {20763--20786},
  year = 	 {2023},
  editor = 	 {Krause, Andreas and Brunskill, Emma and Cho, Kyunghyun and Engelhardt, Barbara and Sabato, Sivan and Scarlett, Jonathan},
  volume = 	 {202},
  series = 	 {Proceedings of Machine Learning Research},
  month = 	 {23--29 Jul},
  publisher =    {PMLR},
  url = 	 {https://proceedings.mlr.press/v202/liang23g.html},
}

@inproceedings{xue2024toward,
  title={Toward effective protection against diffusion-based mimicry through score distillation},
  author={Xue, Haotian and Liang, Chumeng and Wu, Xiaoyu and Chen, Yongxin},
  booktitle={International Conference on Learning Representations},
  volume={2024},
  pages={25734--25758},
  year={2024}
}

@article{salman2023photoguard,
  title={Raising the cost of malicious ai-powered image editing},
  author={Salman, Hadi and Khaddaj, Alaa and Leclerc, Guillaume and Ilyas, Andrew and Madry, Aleksander},
  journal={arXiv preprint arXiv:2302.06588},
  year={2023}
}

@inproceedings{huang2024vbench,
  title={Vbench: Comprehensive benchmark suite for video generative models},
  author={Huang, Ziqi and He, Yinan and Yu, Jiashuo and Zhang, Fan and Si, Chenyang and Jiang, Yuming and Zhang, Yuanhan and Wu, Tianxing and Jin, Qingyang and Chanpaisit, Nattapol and others},
  booktitle={2024 IEEE/CVF Conference on Computer Vision and Pattern Recognition (CVPR)},
  pages={21807--21818},
  year={2024},
  organization={IEEE}
}

@inproceedings{yu2023celebv,
  title={Celebv-text: A large-scale facial text-video dataset},
  author={Yu, Jianhui and Zhu, Hao and Jiang, Liming and Loy, Chen Change and Cai, Weidong and Wu, Wayne},
  booktitle={2023 IEEE/CVF Conference on Computer Vision and Pattern Recognition (CVPR)},
  pages={14805--14814},
  year={2023},
  organization={IEEE}
}

@article{carreira2019kinetics,
  title={A short note on the kinetics-700 human action dataset},
  author={Carreira, Joao and Noland, Eric and Hillier, Chloe and Zisserman, Andrew},
  journal={arXiv preprint arXiv:1907.06987},
  year={2019}
}

@inproceedings{van2023antidb,
  title={Anti-dreambooth: Protecting users from personalized text-to-image synthesis},
  author={Van Le, Thanh and Phung, Hao and Nguyen, Thuan Hoang and Dao, Quan and Tran, Ngoc N and Tran, Anh},
  booktitle={2023 IEEE/CVF International Conference on Computer Vision (ICCV)},
  pages={2116--2127},
  year={2023},
  organization={IEEE}
}

@article{vu2026antii2v,
  title={Anti-I2V: Safeguarding your photos from malicious image-to-video generation},
  author={Vu, Duc and Nguyen, Anh and Tran, Chi and Tran, Anh},
  journal={arXiv preprint arXiv:2603.24570},
  year={2026}
}

@inproceedings{schroff2015facenet,
  title={Facenet: A unified embedding for face recognition and clustering},
  author={Schroff, Florian and Kalenichenko, Dmitry and Philbin, James},
  booktitle={Proceedings of the IEEE conference on computer vision and pattern recognition},
  pages={815--823},
  year={2015}
}

@inproceedings{dong2018boosting,
  title={Boosting adversarial attacks with momentum},
  author={Dong, Yinpeng and Liao, Fangzhou and Pang, Tianyu and Su, Hang and Zhu, Jun and Hu, Xiaolin and Li, Jianguo},
  booktitle={2018 IEEE/CVF Conference on Computer Vision and Pattern Recognition},
  pages={9185--9193},
  year={2018},
  organization={IEEE}
}

@inproceedings{kim2023dense,
  title={Dense text-to-image generation with attention modulation},
  author={Kim, Yunji and Lee, Jiyoung and Kim, Jin-Hwa and Ha, Jung-Woo and Zhu, Jun-Yan},
  booktitle={2023 IEEE/CVF International Conference on Computer Vision (ICCV)},
  pages={7667--7677},
  year={2023},
  organization={IEEE}
}

@article{bai2025qwen3,
  title={Qwen3-vl technical report},
  author={Bai, Shuai and Cai, Yuxuan and Chen, Ruizhe and Chen, Keqin and Chen, Xionghui and Cheng, Zesen and Deng, Lianghao and Ding, Wei and Gao, Chang and Ge, Chunjiang and others},
  journal={arXiv preprint arXiv:2511.21631},
  year={2025}
}

@inproceedings{
wu2024qalign,
title={Q-Align: Teaching {LMM}s for Visual Scoring via Discrete Text-Defined Levels},
author={Haoning Wu and Zicheng Zhang and Weixia Zhang and Chaofeng Chen and Liang Liao and Chunyi Li and Yixuan Gao and Annan Wang and Erli Zhang and Wenxiu Sun and Qiong Yan and Xiongkuo Min and Guangtao Zhai and Weisi Lin},
booktitle={Forty-first International Conference on Machine Learning},
year={2024},
url={https://openreview.net/forum?id=PHjkVjR78A}
}
\bibliographystyle{iclr2027_conference}

\end{document}